\PassOptionsToPackage{table, prologue}{xcolor}

\documentclass[11pt]{article}

\usepackage[]{acl}

\usepackage{times}
\usepackage{latexsym}

\usepackage[T1]{fontenc}

\usepackage[utf8]{inputenc}

\usepackage{amsmath,amssymb}
\usepackage{comment}

\usepackage{xspace}
\usepackage{booktabs}
\usepackage{multicol}
\usepackage{amsmath}
\usepackage{multirow}
\usepackage{graphicx}
\usepackage{enumitem}
\usepackage{mdwlist}
\usepackage{textcomp}
\usepackage{siunitx}
\usepackage{longtable}
\usepackage{verbatim}

\newcommand{\sref}[1]{\S\ref{#1}}
\newcommand{\fref}[1]{Figure~\ref{#1}}
\newcommand{\tref}[1]{Table~\ref{#1}}

\usepackage{microtype}

\title{Your spouse needs professional help: Determining the Contextual  Appropriateness of Messages through Modeling Social Relationships}

\author{
   David Jurgens$^\Diamond$ \\
   University of Michigan \\
   \texttt{jurgens@umich.edu} \\\And
   Agrima Seth$^\Diamond$ \\
   University of Michigan \\
   \texttt{agrima@umich.edu} \\\AND
   Jackson Sargent$^\star$ \\
   University of Michigan \\
  \texttt{jacsarge@umich.edu}\\\And
   Athena Aghighi$^\star$ \\
   University of California, Davis \\
   \texttt{aaghighi@ucdavis.edu}\\\And
   Michael Geraci$^\star$ \\
   University of Buffalo \\
   \texttt{megeraci@buffalo.edu}\\
\\}

\begin{document}
\maketitle
\begin{abstract}

Understanding interpersonal communication requires, in part, understanding the social context and norms in which a message is said. However, current methods for identifying offensive content in such communication largely operate  independent of context, with only a few approaches considering community norms or prior conversation as context. Here, we introduce a new approach to identifying inappropriate communication by explicitly modeling the social relationship between the individuals. We introduce a new dataset of contextually-situated judgments of appropriateness and show that large language models can readily incorporate relationship information to accurately identify appropriateness in a given context.  Using data from online conversations and movie dialogues, we provide insight into how the relationships themselves function as implicit norms and quantify the degree to which context-sensitivity is needed in different conversation settings. Further, we also demonstrate that contextual-appropriateness judgments are predictive of other social factors expressed in language such as condescension and politeness.

\end{abstract}

\section{Introduction}

\def\thefootnote{$\Diamond$}\footnotetext{These authors contributed equally to this work}
\def\thefootnote{$\star$}\footnotetext{These authors contributed equally to this work}
\def\thefootnote{\arabic{footnote}}

\begin{figure}
    \centering
    \includegraphics[width=0.9\linewidth]{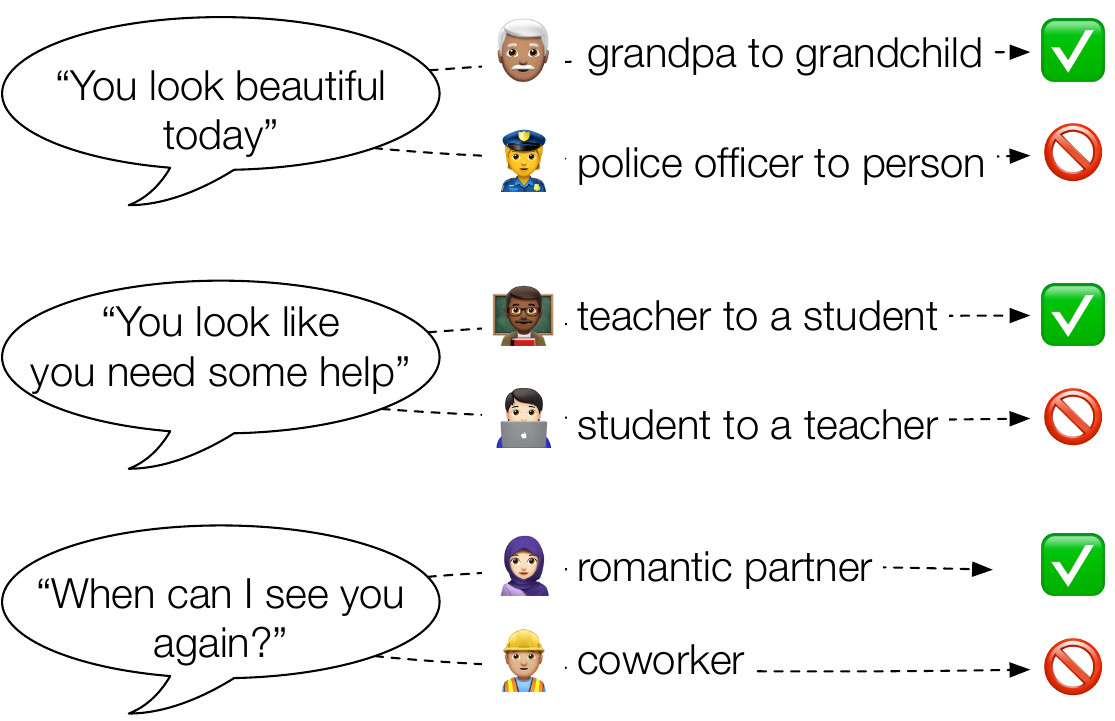}
    \caption{The same message can be appropriate or not depending on the social context in which it is said.}
    \label{fig:intro-example}
\end{figure}

Interpersonal communication relies on shared expectations of the norms of communication \citep{hymes1972communicative}.  
Some of these norms are widely shared across social contexts, e.g., racial epithets are taboo, enabling NLP models to readily identify certain forms of offensive language \citep{fortuna2018survey}. Yet, not all norms are widely shared; the same message said in two different  social contexts may have different levels of acceptability (Figure~\ref{fig:intro-example}). While NLP has recognized the role of social context as important \citep{hovy2021importance,sheth2022defining}, few works have directly incorporated this context into modeling whether messages violate social norms. Here, we explicitly model \textit{relationships} as the social context in which a message is said in order to assess whether the message is appropriate.

NLP models have grown more sophisticated in modeling the social norms needed to identify offensive content. Prior work has shown the benefits of modeling context \citep{menini2021abuse}, such as the demographics of annotators and readers \citep{sap2019risk,akhtar2021whose} and the online community in which a message is said \citep{chandrasekharan2018internet,park2021detecting}. However, these works overlook normative expectations within people's relationships.

In this paper, we introduce a new dataset of over 12,236 instances labeled for whether the message was appropriate in a given relationship context. Using this data, we show that computation models can accurately identify the contextual appropriateness of a message, with the best-performing model attaining a {0.70} Binary F1. Analyzing the judgments of this classifier reveals the structure of the shared norms between relationships. 
Through examining a large corpus of relationship-labeled conversations, we find that roughly 19\% of appropriate messages  could be perceived as inappropriate in another context,  highlighting the need for models that explicitly incorporate relationships.
Finally, we show that our model's relationship-appropriate judgments provide useful features for identifying  subtly offensive language, such as condescension.

\section{Social Norms of Appropriateness}

Relationships are the foundation of society: most human behaviors
and interactions happen within the context of interpersonal relationships \citep{reis2000relationship}. 
Communication norms vary widely across relationships, based on the speakers' social distance, status/power, solidarity, and perceived mutual benefit \citep{argyle1985rules,fiske1992four}.
These norms influence communication in content, grammar, framing, and style \citep{eckert2012constructing} and help reinforce (or subvert) the relationship between speakers \cite{brown1987politeness}.
Prior computational work mostly frames appropriateness as exhibiting positive affect and overlooks the fact that, in some relationships, conversations can be affectively negative but still appropriate \citep{king1984conversational}.
For example, swearing is often considered a norm violation \citep{jay2008pragmatics}, but can also be viewed as a signal of solidarity between close friends \citep{montagu2001anatomy} or co-workers \citep{baruch2007swearing}. 
In such cases, the violation of taboo reinforces social ties by forming a sense of in-group membership where norms allow such messages \citep{coupland2003transgression}.

In sociolinguistics, appropriateness is a function of both context and speech. 
\citet{trudgill1997acts} argues that ``different situations, different topics, different genres require different linguistic styles and registers,'' and \citet{hymes1997scope} argues that the extent to which ``something is suitable, effective or liked in some context'' determines its appropriateness. Whether a discourse is appropriate depends strongly on the social context in which it is produced and received \citep{fetzer2015appropriateness}, making the assessment of appropriateness a challenging task due to the need to explicitly model contextual norms. Behavioral choices are subject to the norms of ``oughtness'' \citep{harre1972explanation,shimanoff1980communication}, and \citet{floyd1997affectionate} suggest relationship types as an important factor influencing the normative expectations for relational communication. For example, while it may be considered appropriate for siblings to discuss their past romantic relationships in detail, the topic is likely to be perceived as taboo or inappropriate between romantic partners \citep{baxter1985taboo}.

\section{Building a Dataset of Contextual Appropriateness}

Prior work has shown that interpersonal relationships are a relevant context for the appropriateness of content \citep{locher2010interpersonal}. 
While not all messages differ in this judgment---e.g., ``hello'' may be appropriate in nearly all settings---building a dataset that embodies this context sensitivity remains a challenge. 
Here, we describe our effort to build a new, large dataset of messages rated for contextual appropriateness, including how we select relationships and operationalize appropriateness. Due to the challenge of identifying and rating these messages, our dataset is built in two phases. 

\paragraph{Selecting Relationships}

Formally categorizing relationships has long been a challenging task for scholars \citep{regan2011close}. 
We initially developed a broad list of relationships, drawing from 1) folk taxonomies~\citep{berscheild89rci}, e.g., common relationship types of friends \citep{adams2000definitions}, family \citep{gough1971origin}, or romantic partners \citep{miller2007intimate}; and 2) organizational and social roles \citep{stamper2009typology}, e.g., those in a workplace, classroom, or functional settings, as these frequently indicate different social status, distance, or solidarity between individuals in the relationship. Using this preliminary list, four annotators performed a pilot assessment of coverage by discussing quotes from movie scripts, social media, or their imagination and identifying cases where an excluded relationship would have a different judgment for appropriateness. Ultimately, 49 types of relationships were included, shown in Table \ref{tab:relationships}. 

\begin{table}[t]
 \resizebox{\linewidth}{!}{
\rowcolors{2}{gray!15}{white}
\begin{tabular}{ c p{0.83\linewidth}} 

\textbf{Category} & \textbf{Relationships} \\
\hline

\textsc{Family} & parent,$^\dagger$ child,$^\dagger$ adopted child,$^\dagger$ siblings, step-siblings, grandparent,$^\dagger$ grandchild,$^\dagger$ niece/nephew,$^\dagger$ cousins, uncle/aunt$^\dagger$   \\

\textsc{Social} &  best friend, friend, old friend, childhood friend, acquaintance, neighbor, complete stranger  \\

\textsc{Romance}  &  dating, engaged, married, domestic partner, friends with benefits, a person whom one has an affair with, divorcee, ex-boyfriend/ex-girlfriend \\

\textsc{Organizational}  & coworker, colleague, another employee in a larger company, boss$^\dagger$ (to a direct report), direct report$^\dagger$ (to a boss) \\ 

\textsc{Peer Group} & classmate, sports teammate, club member\\ 

\textsc{Parasocial} &  fan,$^\dagger$ hero$^\dagger$ \\

\textsc{Role-based} & law enforcement,$^\dagger$ individual with authority$^\dagger$ (generic), mentor,$^\dagger$ mentee,$^\dagger$ teacher,$^\dagger$ student,$^\dagger$ lawyer,$^\dagger$ client,$^\dagger$ doctor,$^\dagger$ patient,$^\dagger$ landlord$^\dagger$  \\

\textsc{Antagonist} & competitor, rival, enemy \\ 
\end{tabular}
}
\caption{The organization of relationships into folk categories. Relationships with asymmetric reciprocal roles are marked with a $^\dagger$, e.g., parent and child. In the context of annotation, these relationships are interpreted as being spoken from that role to the reciprocal role, e.g., the parent is interpreted as ``parent to a child'' and the doctor is ``doctor to a patient.'' }
\label{tab:relationships}
\end{table}

\paragraph{Defining Appropriateness}

Appropriateness is a complex construct that loads on many social norms \cite{fetzer2015appropriateness}. For instance, in some relationships, an individual may freely violate topical taboos, while in other relationships, appropriateness depends on factors like deference due to social status. Informed by the theory of appropriateness~\citep{march2004logic}, we operationalize \textit{inappropriate} communication as follows: 
Given two people in a specified relationship and a message that is plausibly said under normal circumstances in this relationship, would the listener feel offended or uncomfortable?
We use plausibility to avoid judging appropriateness for messages that would likely never be said, e.g., ``would you cook me a hamburger?'' would not be said from a doctor to a patient. We constrain the setting to what an annotator would consider normal circumstances for people in such a relationship when deciding whether the message would be perceived as appropriate; for example, having a teacher ask a student to say something offensive would be an \textit{abnormal} context in which that message is appropriate. Thus, during annotation, annotators were asked to first judge if the message would be plausibly said and only, if so, rate its appropriateness.

Judging appropriateness necessarily builds on the experiences and backgrounds of annotators. Culture, age, gender, and many other factors likely influence decisions on the situational appropriateness of specific messages. In making judgments, annotators were asked to use their own views and not to ascribe to a judgment of a specific identity. 

\paragraph{Raw Data}

Initial conversational data was selectively sampled from English-language Reddit. Much of Reddit is not conversational in the sense that comments are unlikely to match chit-chat. Further, few comments are likely to be context-sensitive. To address these concerns, we filter Reddit comments in two ways. First, we train a classifier to identify conversational comments, using 70,949 turns from the Empathetic dialogs data~\citep{rashkin2019empathetic} and 225,907 turns from the Cornell movie dataset~\citep{danescu2011chameleons} as positive examples of conversational messages, and 296,854 turns from a random sample of Reddit comments as non-conversational messages. Full details are provided in Appendix \ref{app:conversational}. Second, we apply our conversational classifier to comments marked by Reddit as controversial in the Pushshift data~\citep{baumgartner2020pushshift}; while the decision logic for which comments are marked as controversial is proprietary to Reddit, controversial-labeled comments typically receive high numbers of both upvotes and downvotes by the community---but are not necessarily offensive. These two filters were applied to identify 145,210 total comments gathered from an arbitrary month of data (Feb. 2018).

\subsection{Annotation Phase 1}

In the first phase of annotation, four annotators individually generated English-language messages they found to differ in appropriateness by relationship.\footnote{This process was developed after pilot tests showed the random sample approach was unlikely to surface interesting cases, but annotators found it easier to ideate and write their own messages after being exposed to some example communication.} Annotators were provided with a website interface that would randomly sample conversational, controversial Reddit comments as inspiration. Details of the annotation instructions and interface are provided in Appendix \ref{app:annotation}. The annotation process used a small number of in-person annotators rather than crowdsourcing to allow for task refinement: During the initial period of annotating, annotators met regularly to discuss their appropriateness judgments and disagreements. %
This discussion process was highly beneficial for refining the process for disentangling implausibility from inappropriateness.
Once annotation was completed, annotators discussed and adjudicated their ratings for all messages. Annotators ultimately produced 401 messages and 5,029 total appropriateness ratings for those messages in the context of different relationships.

\begin{table*}[tb]

\resizebox{\textwidth}{!}{
\rowcolors{2}{gray!15}{white}
\begin{tabular}{p{0.4\textwidth} p{0.4\textwidth} p{0.4\textwidth}  }
\toprule
Message & Appropriate Relationship Contexts & Inappropriate Relationship Contexts \\
\midrule
You're so grown up now! & grandparent, cousins, neighbor, parent, uncle aunt & direct report (to a boss), student (to a teacher) \\
Sorry, were we 2-0 against you?  I forget. & rival, competitor & club member, sports teammate \\
Pull your car over! & law enforcement & complete stranger, competitor \\
You need to get out more. & friend, domestic partner, sibling, best friend, parent & complete stranger \\

She is actually so attractive. 
& sibling, grandchild (to grandparent), domestic partner, to a person one is dating, childhood friend, child, adopted child, best friend, classmate, parent, friend 
& colleague, boss, teacher (to a student), student (to a teacher), mentor, mentee (to mentor),   direct report (to a boss), law enforcement, co-worker \\

I'm afraid you're right. But it's also time to move on 
& teacher, boss, colleague, sibling, domestic partner, childhood friend, ex-lover,  %
& mentee (to a mentor), direct report (to a boss), student (to a teacher) \\

So how was the date last night bro & dating, best friend, friend & law enforcement, direct report, person with authority, employee in large company \\

I'm glad we're friends & friend, sibling, childhood friend, old friend, cousins, best friend, step sibling & complete stranger, acquaintance, friends with benefits \\

How would you know? 
& colleague, boss, sibling, lawyer (to client), doctor, best friend, classmate, step sibling, friend, dating, law enforcement  
& mentee (to mentor), complete stranger, teacher (to a student), patient (to doctor), child (to parent),  parent (to child),neighbor, spouse, old friend,  \\

Oh I see, I'm sorry I misunderstood 
& colleague, teacher, student, lawyer, boss, sibling, club member, grandchild, doctor, complete stranger, %
&  \\

\bottomrule
\end{tabular}
}
\caption{Examples of the labeled data with a sample of the relationship contexts that annotators viewed as being appropriate or not for the message. }
\label{tab:examples}
\end{table*}

\subsection{Annotation Phase 2}
\label{sec:phase2}

Phase 2 uses an active learning approach to identify potentially relationship-sensitive messages to annotate from a large unlabeled corpus. A T5 prompt-based classifier was trained using OpenPrompt \citep{ding2022openprompt} to identify whether a given message would be appropriate to say to a person in a specific relationship. Details of this classifier are provided in Appendix \ref{app:pilot-classifier}. This classifier was run on all sampled data to identify instances where at least 30\% of relationships were marked as appropriate or inappropriate; this filtering biases the data away from universally-appropriate or inappropriate messages, though annotators may still decide otherwise. 

Two annotators, one of which was not present in the previous annotation process, completed two rounds of norm-setting and pilot annotations to discuss judgments. Then, annotators rated 30 messages each, marking each for plausibility and, if plausible, appropriateness; they met to adjudicate and then rated another 41 messages. This produced 2,159 appropriateness ratings across these messages. Annotators had a Krippendorff's $\alpha$ of 0.56 on plausibility and, for messages where both rated as plausible, 0.46 on appropriateness. While this agreement initially seems moderate, annotators reviewed all disagreements, many of which were due to different interpretations of the same message, which influenced appropriate judgments rather than disagreements in appropriateness itself. Annotators then revised their own annotations in light of consensus in message meaning, bringing the plausibility agreement to 0.72 and appropriateness to 0.92. We view these numbers as more reliable estimates of the annotation process, recognizing that some messages may have different judgments due to annotators' values and personal experiences. We mark the 2,159 ratings in this data as \textit{Adjudicated} data for later evaluation.

Both annotators then independently annotated different samples of the Reddit data in order to maximize diversity in messages. Annotators were instructed to skip annotating messages that they viewed as less context-sensitive (e.g., offensive in all relationship contexts) or where the message did not appear conversational. Annotators provided 5,408 ratings on this second sample. We refer to this non-adjudicated data as Phase 2 data.

\subsection{Dataset Summary and Analysis}

The two phases produced a total of 12,236 appropriateness judgments across 5299 messages. Of these, 7,589 of the judgments were appropriate, and 4647 were inappropriate. Table~\ref{tab:examples} shows examples of annotation judgments.
In line with prior cultural studies of appropriateness \citep{floyd1997affectionate,fetzer2015appropriateness}, three themes emerged during training. 
First, annotators noted the perception of the role of \textit{teasing} in deciding appropriateness. Teasing messages are directed insults (mild or otherwise) aimed at the other party; comments such as ``you are so dumb'' are likely made in jest within close relationships such as best friends or siblings but inappropriate in many others. 
Second, messages' appropriateness depended in part on whether the relationship was perceived to be supportive; for example, the message ``At least you called him by his correct name'' could be one of encouragement in the face of a mistake (e.g., if said by a spouse) or a subtle insult that implies the listener \textit{should have} known more about the third party. 
Third, differences in the power/status in the relationship influenced appropriateness, where very direct messages, e.g., ``you made a mistake there.'' were often perceived to be inappropriate when said to a person of higher status, a known violation of politeness strategies \citep{brown1987politeness}.
Ultimately, appropriateness was judged through a combination of these aspects.

\begin{figure}
  \centering
  \includegraphics[width=\linewidth]{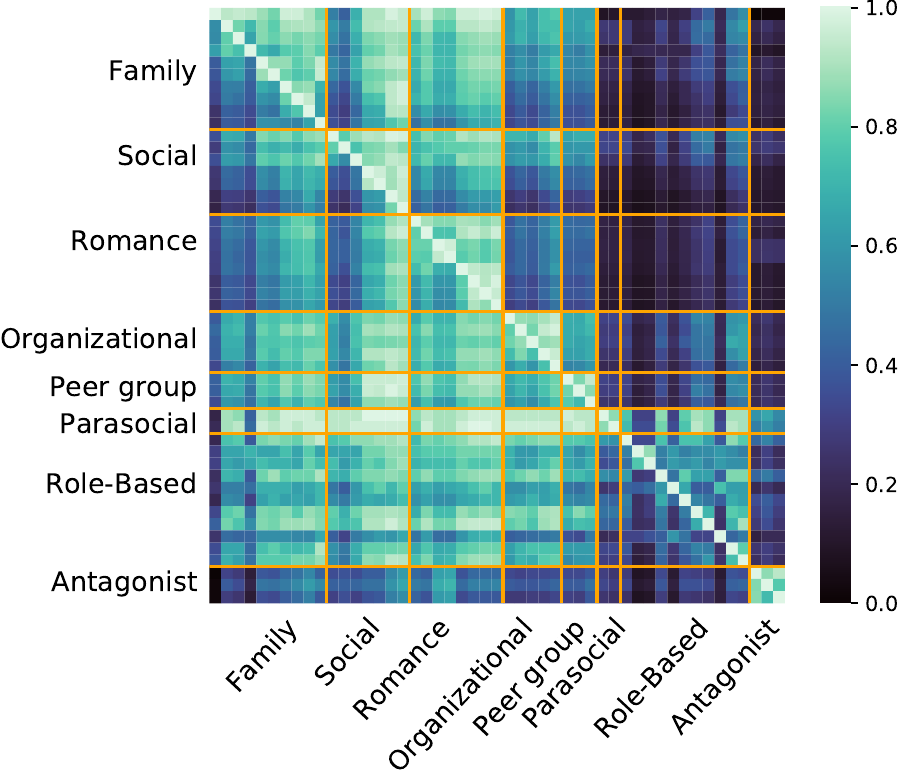}
  \caption{The probability that, given an appropriate message for the relationships represented by a row, the message will also be appropriate in another relationship listed in the column. Probabilities are calculated across the entire  data.}
  \label{fig:app-heatmap}
\end{figure}

As an initial test of regularity in how the relationship influence perceived appropriateness, we measured the probability that a message appropriate for relationship $r_i$ is also appropriate for $r_j$ using all the annotations, shown in \fref{fig:app-heatmap} and grouped by thematic categories. Clear block structure exists with some categories, e.g., \textsc{Organization}, indicating shared norms of appropriateness for relationships within the same category. %
In contrast, the \textsc{Family} and \textsc{Social} categories contain relationships with different power (e.g., parent) and social distance (e.g., friend vs. stranger), leading to varied judgments. Figure \ref{fig:app-heatmap} also reveals the  asymmetry in which message themes are appropriate: While much of what is said for \textsc{Role-based} relationships is also appropriate in \textsc{Social} or \textsc{Romance}, the reverse is not true.

\section{Identifying Contextual Inappropriateness}
\label{sec:identifying}
Given the high potential agreement of annotators, we test whether models can similarly recognize whether a given message is appropriate if said in the context of a specific relationship.

\paragraph{Experimental Setup}

Two classes of models were trained. Given the recent successes of prompt-based models, we build models using the OpenPrompt library~\citep{ding2022openprompt} and, to support larger models, using the PEFT library~\citep{liu2022few}.
The OpenPrompt library was used to train \texttt{t5-base} and \texttt{gpt2-med} models using the prompt {``Is it appropriate for {person1} to say {"quote"} to {person2}, "yes" or "no"? \texttt{[MASK]}'' using the verbalization ``yes' or ''no'' for the masked token}. Here, we fill in {person1} and {person2} to refer to the two parties in the relationship. Examples of filled-in templates and other prompts are reported in Appendix~\ref{app:prompt-models}, though performance varied only slightly across different prompts. 
The PEFT library was used to train the \texttt{large} and \texttt{xl} variants of the \texttt{flan-t5} model \citep{chung2022scaling}. This model has been pretrained for instruction following; therefore, based on suggested guidelines from \citet{ziems2023can}, we use the following prompt: ``Rate whether it is inappropriate for this message to be said in the following social setting?\textbackslash n setting: {relationship description} \textbackslash n message: {"quote"}\textbackslash n answer (yes or no):'' Due to the resources required for training these larger models, no additional prompts were rigorously evaluated outside of initial pilot testing.

The second class of models uses masked language model (MLM) fine-tuning on the \texttt{[CLS]} token from an MLM to predict appropriateness. Here, we frame the instance using the same language as the OpenPrompt-based models but fill in the MASK with ``yes'' (i.e., indicating that the message is appropriate to say in the relationship). The classification model is then fine-tuned to classify whether this hard-coded judgment is correct or not. We test two recent MLMs, MiniLM \cite{wang2020minilm}, a small distilled model, and DeBERTa-v3 \cite{he2021debertav3}, a much larger model. These two models reflect extremes among relatively small MLMs and allow us to assess whether more social relationship knowledge might be embedded within a larger parameter space.

Annotated data was split at the message level 70:10:20 into train, development, and test sets, resulting in 9,107 train, 1,100 development, and 2,029 test instances. We frame the task similar to offensive language detection and use Binary F1 as our metric where inappropriate is the positive class. Model performance is reported as the average across five random runs. Additional training details and per-seed performance are provided for all systems in Appendix~\ref{app:more-results}.

Two baseline systems are included. The first is random labels with respect to the empirical distribution in the training data. The second uses Perspective API \citep{lees2022new} to rate the toxicity of the message, labeling it as toxic if the rating is above 0.7 on a scale of [0,1]; the same label is used for all relationships. While this baseline is unlikely to perform well, it serves as a reference to how much explicit toxicity is in the dataset, as some (though not all) of these messages are inappropriate to all relationships.

\paragraph{Results}

Models accurately recognized how relationships influence the acceptability of a message, as seen in \tref{tab:model-performance}. Prompt-based models were largely equivalent to MLM-based models, though both approaches far exceeded the baselines. The largest model, \texttt{flan-t5-xl}, ultimately performed best, though even the MiniLM offered promising performance, despite having several orders of magnitude fewer parameters.  In general, models were more likely to label messages as inappropriate even when appropriate for a particular setting (more false positives). This performance may be more useful in settings where a model flags potentially inappropriate messages which are then reviewed by a human (e.g., content moderation). However, the performance for models as a whole suggests there is substantial room for improvement in how relationships as social context are integrated into the model's decisions.

\begin{table}[t!]
 \resizebox{0.48\textwidth}{!}{
 \rowcolors{2}{blue!05}{white}
\begin{tabular}{ c c ccc } 

 \textbf{Model} & \textbf{Type} & \textbf{Precision} & \textbf{Recall} & \textbf{F1} \\
\hline
 \textit{random} & \textit{n/a} & 0.436 & 0.368 & 0.399  \\
 \textit{Perspective API} & \textit{n/a} & 0.422 & 0.097 & 0.157 \\
 \texttt{DeBERTa-v3}      & MLM fine-tuning   & {0.658} & {0.660} & {0.659}\\
 \texttt{MiniLM}         & MLM fine-tuning   & {0.615} & {0.705} & {0.656}\\
 \texttt{t5-base}         & prompt-based     & {0.655} & {0.683} & {0.669}\\ 
 \texttt{gpt2-med}        & prompt-based     & {\textbf{0.668}} & {0.650} & {0.665}\\ 
 \texttt{flan-T5-large}   & prompt-based     & {0.626} & {0.704} & {0.661}\\
 \texttt{flan-t5-xl}      & prompt-based     & {0.666} & {\textbf{0.736}} & {\textbf{0.698}}\\

\end{tabular}
}
\caption{Performance (Binary F1) at recognizing whether a message was  \textit{in}appropriate in a relationship context.}
\label{tab:model-performance}
\end{table}

\paragraph{Error Analysis}

\begin{figure}[t]
    \centering
    \includegraphics[width=\linewidth]{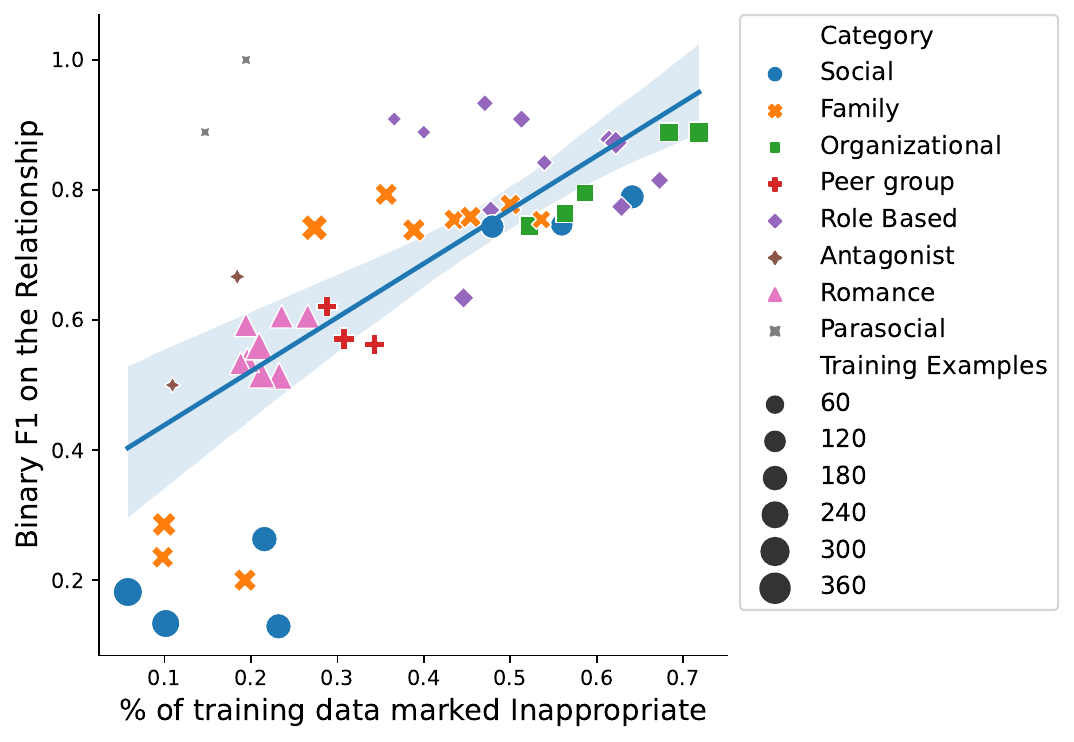}
    \caption{ Performance of the \texttt{flan-t5-xl} model relative to the \% of training examples marked inappropriate per relationship. Colors denote the relationship category and sizes the number of training examples.  }
    \label{fig:error-analysis}
\end{figure}

Different relationships can have very different norms in terms of what content is acceptable, as highlighted in \fref{fig:app-heatmap}. How did model performance vary by relationship? Figure \ref{fig:error-analysis} shows the binary F1 score of the \texttt{flan-t5-xl} model by relationship, relative to the percent of training instances the model saw that were inappropriate; Appendix Table~\ref{tab:per-rel-performance} shows full results per relationship.
Model performance was highly correlated with the data bias for inappropriateness ($r$=0.69; p<0.01). The model had trouble identifying inappropriate comments for relationships where most messages are appropriate (e.g., friend, sibling) in contrast to more content-constrained relationships (boss, student, doctor). These low-performance relationships frequently come with complex social norms---e.g., the boundary between appropriate teasing and inappropriate hurtful comments for siblings \citep{keltner2001just}---and although such relationships have among the most training data, we speculate that additional training data is needed to model these norms, especially given the topical diversity in these relationships' conversations.

\section{Generalizing to Unseen Relationships}
\label{sec:ablation}

Through their pretraining, LLMs have learned semantic representations of relationships as tokens. Our classification experiments show that LLMs can interpret these relationship-as-token representations to effectively judge whether a message is appropriate. To what extent do these representations allow the model to generalize about new relationships not seen in training? In particular, are models able to generalize if a category of relationship, e.g., all family relations, was never seen? Here, we conduct an ablation study where one of our folk categories is held out during training.

\paragraph{Setup}

The \texttt{flan-t5-xl} model is trained with the same hyperparameters as the best-performing system on the full training data. We use the same data splits, holding out all training examples of relationships in one category during training. We report the Binary F1 from the test set on (1) relationships seen in training and (2) relationships in the held-out category. Note that because training set sizes may change substantially due to an imbalance of which relationships were annotated and because  categories have related norms of acceptability, performance on seen-in-training is likely to differ from the full data.

\paragraph{Results}

\begin{figure}[t]
    \centering
    \includegraphics[width=\linewidth]{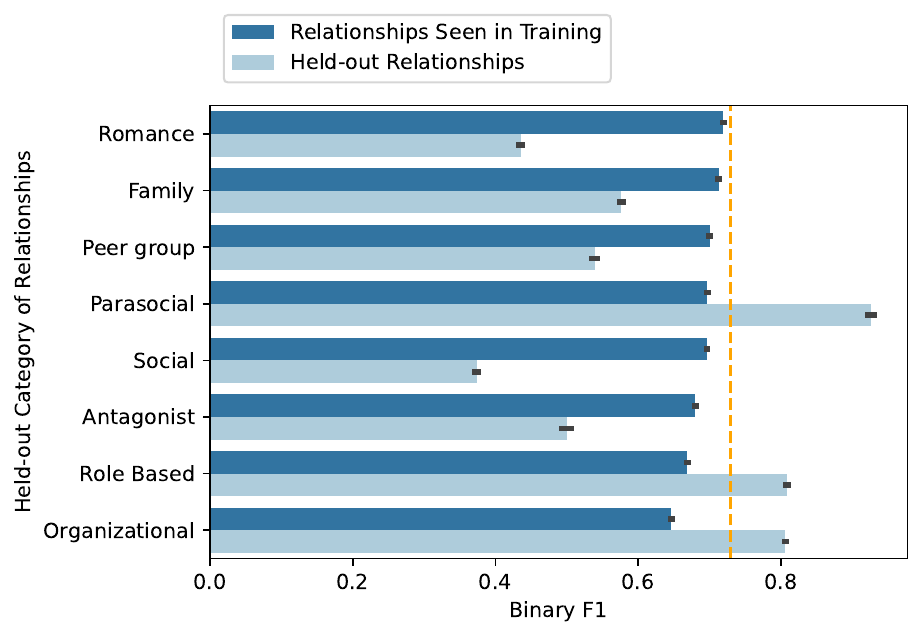}
    \caption{Performance by the T5 Prompt-based model at rating contextual appropriateness to relationships seen in training versus categories of unseen relationships. Ablated models are ordered by performance on seen relationships with the vertical dashed line marking when training on all data on the test set; error bars show 95\% bootstrapped confidence intervals. }
    \label{fig:ablation}
\end{figure}

Ablated models varied substantially in their abilities to generalize to the unseen relationship types, as well as in their baseline performance (Figure \ref{fig:ablation}). 
First, when ablating the larger categories of common relationships (e.g., \textsc{Family}, \textsc{Social}), the model performs well on seen-relationships, dropping performance only slightly, but is unable to accurately generalize to relationships in the unseen category.
These unseen categories contain relationships that span a diverse range of norms with respect to power differences, social distance, and solidarity. While other categories contain partially-analogous relationships along these axes, e.g., parent-child and teacher-student both share a power difference, the drop in performance on held-out categories suggests the model is not representing these social norms in a way that allows easy transfer to predicting appropriateness for unseen relationships with similar norms.
Second, relationships in three categories improve in performance when unseen: \textsc{Organizational}, \textsc{Role-Based}, and \textsc{Parasocial}. All three categories feature relationships that are more topically constrained around particular situations and settings. While the categories do contain nuance, e.g., the appropriateness around the power dynamics of boss-employee, the results suggest that models may do well in zero-shot settings where there is strong topic-relationship affinity---and messages outside of normal topics are inappropriate.
Viewing these two trends together, we posit that the semantic representations of relationships in \texttt{flan-t5-xl} currently capture only minimal kinds of social norms---particularly those relating to topic---and these norms are not represented in a way that lets the model easily generalize to reasoning about relationships not seen in training.

\section{How Much of Conversation is Context Sensitive in Appropriateness?}
\label{sec:pride}

Our annotation and computational models have shown that the relationship context matters in determining appropriateness. However, it is unclear how often conversations are sensitive to this context. For example, the majority of conversation may be appropriate to all relationships. Here, we aim to estimate this context sensitivity by testing the appropriateness of a message in counterfactual settings using an existing dataset labeled with relationship types.

\paragraph{Experimental Setup}

To estimate context sensitivity, we use our most accurate model to label a large selection of dialog turns from the PRIDE dataset \citep{tigunova-etal-2021-pride}. PRIDE consists of 64,844 dialog turns from movie scripts, each annotated for the relationship between the speaker and receiver, making it ideal as a high-plausibility conversational message said in relationships. However, some turns of the dialog are explicitly grounded in the setting of the movie, e.g., ``How's it going, Pat?'' which makes the turn too specific to that particular setting to accurately estimate appropriateness. Therefore, we run SpaCy NER \cite{spacy2} on the dialog and remove all turns containing references to people, companies, countries, and nationalities in order to keep the dialog generic and maximally plausible in many different relationship contexts. Further, we remove turns with only a single token or over {100} tokens. This filtering leaves  47,801 messages for analysis. 

PRIDE contains 18 unique relationships, 16 of which were already included in our categories (cf. \tref{tab:relationships}); the two previously-unseen relationship types, described as ``religious relationships'' and ``client/seller (commercial),'' were also included since our model can accommodate zero-shot prediction.\footnote{These relationships were phrased as ``from a person to someone in their church'' and ``from a person to a commercial associate'' in our prompt model testing.}

To text for context sensitivity, we apply our \texttt{flan-t5-xl} model and measure the appropriateness of the actual relationship context and then the counterfactual cases as if the message had been said in an alternative relationship context seen in their data.  This setup allows us to assess whether if a message was appropriate in its intended relationship context, would it still be appropriate in another.

\paragraph{Results} 

\begin{figure}[t]
    \centering
    \includegraphics[width=0.48\textwidth]{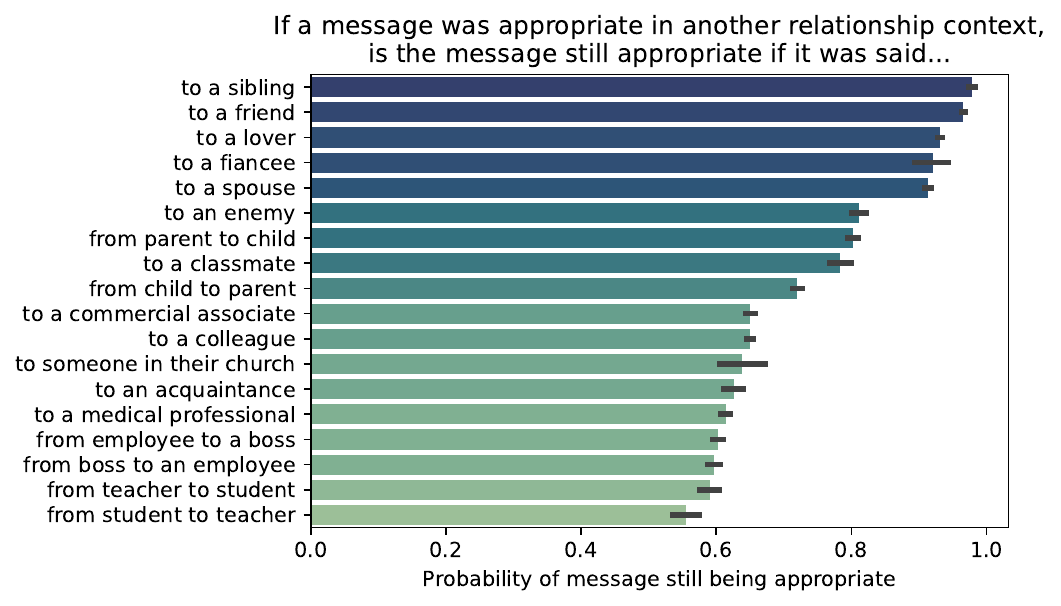}
    \caption{  
    If a message was appropriate to some other relationship, what is the probability that that message would also be appropriate if said in a different relationship context (shown on the y-axis)? The relative differences show a clear separation between close social ties and those relationships with high social distance or large power differences. }
    \label{fig:pride-cs}
\end{figure}

Considering only appropriate messages and excluding the unusual \textit{enemy} relationship from consideration, we find that roughly 19\% of the appropriate-as-said messages in the data would be inappropriate if said in the context of a different relationship. Figure \ref{fig:pride-cs} shows the probability that a message acceptable in some other relationship context would also be acceptable in the given context; the striking decrease in the likelihood of acceptability follows the increasingly constrained social norms around a relationship. For example, while friends and loved ones have broad latitude to discuss sensitive topics~\citep{hays84friendshipdevelopment}, \textsc{Role-based} relationships and those with larger power differences are more constrained in what is considered acceptable conversation. While the movie dialog in the PRIDE dataset likely  differs from a natural dialog, these results point to relationships as important contexts in natural language understanding. 

More generally, we suggest a need for socially-aware models to identify offensive language. While substantial effort has been put into identifying explicit toxic or abusive language~\citep{vidgen2021abuse}, few models, if any, incorporate the context in which the message is said. These models typically rely on previous conversation turns~\citep{zhang2018conversations} or modeling community-level social norms~\citep{chandrasekharan2018internet} to understand how the context may shift whether the message is perceived as appropriate. Our result suggests that the social context---and particularly social relationships---are highly influential in measuring appropriateness. Indeed, together with the result showing the (expected) low performance of the Perspective API toxicity detector, these results suggest NLP models deployed in social settings are likely missing identifying many offensive messages due to their lack of explicitly modeling of social relations. As NLP tools make their way into the workplace setting, which frequently features a mix of \textsc{Organizational}, \textsc{Social}, and \textsc{Romance} ties, explicitly modeling context will likely be necessary.

\section{Identifying Subtle Offensiveness using Contextual Appropriateness}

Prior NLP studies of subtly inappropriate language often omit the social context in which a statement is said \citep{breitfeller2019finding,perez2022semeval}, yet it is often this context that makes a statement inappropriate. For example, a teacher asking a student ``Do you need help writing that?'' is appropriate, whereas a student asking a teacher the same question may seem rude. We hypothesize that modeling the relative appropriateness of a message across relationships can help identify types of subtly offensive language. We test this hypothesis using datasets for two phenomena: condescension \cite{wang2019talkdown} and (im)politeness \cite{danescu2013computational}. 

\paragraph{Experimental Setup}

The \texttt{flan-t5-xl} model is used to predict the appropriateness of each message in the training data in the TalkDown dataset for condescension \cite{wang2019talkdown}, and the Stanford Politeness Corpus \cite{danescu2013computational}. Each message is represented as a binary vector of inappropriateness judgments for each relationship. TalkDown is based on Reddit comments, which our model has seen, whereas the politeness data is drawn from Wikipedia and StackExchange conversations. 
We adopt the same train and test splits as in the respective papers and fit a logistic regression classifier for each dataset to predict whether a message is condescending or impolite, respectively, from the per-relationship appropriateness vector. The logistic regression model uses Scikit-learn~\citep{pedregosa2011scikit}; for each task, we adopt the evaluation metric used in the respective paper.  Appendix~\ref{app:subtle} has additional details.

\paragraph{Results}

The relationship appropriateness scores were meaningfully predictive of subtle offensiveness, as seen in \tref{tab:talkdown} for condescension and \tref{tab:politeness} for impoliteness. In both settings, the appropriateness features provide a statistically significant improvement over random performance, indicating that adding relationships as context can help identify subtly offensive messages.
Further, despite the classifier's relative simplicity, the appropriateness features alone outperform the \texttt{bert-large} classifier used in \citet{wang2019talkdown} in the balanced setting, underscoring how explicitly modeling relationships can still be competitive with LLM-based approaches. 
Performance at recognizing (im)politeness from relationship-appropriateness was lower than the hand-crafted or purely bag-of-words approaches. Yet, this gap is expected given that dataset's design; \citet{danescu2013computational} focus on identifying discourse moves, and the politeness classification task comes from messages at the top and bottom quartiles of their politeness rating. Messages in the bottom quartile may be less polite, rather than impolite, and therefore appropriate in more context, thereby making relationship-appropriate judgments less discriminating as features.

\begin{table}[t]
 \resizebox{0.48\textwidth}{!}{
\begin{tabular}{ ccc } 

 Model & Imbalanced Data & Balanced Data \\
 \hline
 Appropriateness Feats. & {0.624} & {\textbf{0.708}} \\
 \texttt{bert-large}    & \textbf{0.684} & 0.654 \\
 \texttt{bert-base}     & 0.657 & 0.596 \\
 \textit{random}        & 0.371 & 0.500 \\
 \textit{majority}      & 0.488 & 0.333 \\

\end{tabular}
}
\caption{Comparison of performance (macro-F1) for predicting condescension on balanced and imbalanced datasets of TalkDown \citep{wang2019talkdown} using contextual appropriateness ratings as a feature. } %
\label{tab:talkdown}
\end{table}

\begin{table}[t]
 \resizebox{0.48\textwidth}{!}{
\begin{tabular}{ c c c c c } 
 & \multicolumn{2}{c}{In-domain} & \multicolumn{2}{c}{Cross-domain}\\
 \cline{2-5}
 Train & Wiki & SE  & Wiki & SE \\
 Test  & Wiki & SE  & SE & Wiki \\
 \hline
 Appropriateness Feats.  & {69.11} & {57.81} & {57.63}  & {64.86}\\
 Bag of Words            & 79.84 & 74.47 & 64.23 & 72.17\\ 
 Politeness Feats.       & \textbf{83.79} & \textbf{78.19} & \textbf{67.53} & \textbf{75.43}\\   
 \hline
 \textit{Random}   & 49.15 & 48.64 & 51.3  & 48.59 \\
 \textit{Human}    & 86.72 & 80.89 & 80.89 & 86.72\\   

\end{tabular}
}
\caption{Comparison of performance (accuracy) for predicting politeness from contextual appropriateness ratings as  features. 
Data and comparison results are from \citet{danescu2013computational}. }
\label{tab:politeness}
\end{table}

\section{Conclusion}

``Looking beautiful today!'', ``You look like you need a hand with that'', and ``When can I see you again?''---in the right contexts, such messages can bring a smile, but in other contexts, such messages are likely to be viewed as inappropriate. In this paper, we aim to detect such inappropriate messages by explicitly modeling the relationship between people as a social context. Through a large-scale annotation, we introduce a new dataset of over 12,236 ratings of appropriateness for 49 relationships. In experiments, we show that models can accurately identify inappropriateness by making use of pre-trained representations of relationships. 
Further, through counterfactual analysis, we find a substantial minority of content is contextually-sensitive: roughly 19\% of the appropriate messages we analyzed would not be appropriate if said in some other relationship context.
Our work points to a growing need to consider meaning within the social context, particularly for identifying subtly offensive messages. All data and code are released at {\small \url{https://github.com/davidjurgens/contextual-appropriateness}}. 

\section*{Acknowledgments}

The authors thank Aparna Anathasubramaniam, Minje Choi, and Jiaxin Pei for their timely and valuable feedback on the paper. This work was supported by the National Science Foundation under Grant Nos. IIS-1850221, IIS-2007251 and  IIS-2143529. 

\section{Limitations}

This paper has three main limitations worth noting.
First and foremost, while our paper aims to model the social context in which a message is said, the current context is limited to only the parties' relationship. In practice, the social context encompasses a wide variety of other factors, such as the sociodemographics of the parties, the culture and setting of the conversation,  and the history of the parties. Even relationships themselves are often much more nuanced and the appropriateness may vary widely based on setting, e.g., statements said between spouses may vary in appropriateness when made in public versus private settings. These contextual factors are likely necessary for a full account of the effect of social context on how messages should be perceived.
Our work provides an initial step in this direction by making the relationship explicit, but more work remains to be done.
Future work may examine how to incorporate these aspects, such as by directly inputting the situation's social network as context using graph embedding techniques \citep{kulkarni-etal-2021-lmsoc-approach}, where the network is  labeled with relationships \citep{choi2021more}, or by modeling relationships particular types of settings such as in-person, phone, texting, or other online communication, which each have different norms.

Second, our data includes annotations on a finite set of relationships, while many more unique relationships are possible in practice, e.g., customer or pastor. Our initial set was developed based on discussions among annotators and aimed at high but not complete coverage due to the increasing complexity of the annotation task as more relationships were added.
Our results in Section \ref{sec:ablation} suggest that our best model could be able to generalize to new types of relationships in some settings and  zero-shot results on two new relationship types not seen in training (a fellow church member and a commercial relationship) match expectations of context sensitivity, (cf. Figure~\ref{fig:pride-cs} .
However, performance is likely limited for less-common relationships without additional training data to describe the norms of appropriateness in this context; and, based on the error analysis in Section \ref{sec:identifying}, models are currently unlikely to generalize to unseen relationships that have complex sensitivity norms.
In addition, new settings such as online spaces may require additional definitions of relationships as individuals interact with each other anonymously.

Third, our judgments of appropriateness were drawn from five annotators total, each of whom had different views of appropriateness based on their values and life experience. While our analysis of agreement with the Adjudicated data (Section \ref{sec:phase2}) suggests that when annotators can reach a consensus on a message's meaning, they are highly likely to agree on appropriateness, we nonetheless view that our annotations are likely to primarily reflect the values of the annotators and may not generalize to other social or cultural contexts where the norms of relationships differ. Future work is needed to explore how these norms differ through additional annotation, and we hope that our dataset will provide a reference for comparison to these judgments.
For example, future work may make use of annotation schemes that explicitly model disagreements \citep{fornaciari-etal-2021-beyond} or personalized judgments \citep{plepi-etal-2022-unifying}; such approaches may be able to better represent common factors influencing appropriateness judgments.

\section{Ethical Considerations}

We note three points on ethics. First, we recognize that appropriateness is a value judgment, and therefore our data is limited here by the viewpoints of the annotators. Multiple works on offensive language have shown that the values and identities of annotators can bias the judgments and potentially further marginalize communities of practice whose views and norms are not present~\citep{sap2019risk,garg2022handling}. We have attempted to mitigate this risk by adding diversity to our annotator pool with respect to gender, age, and culture, yet our limited  pool size necessitates that not all viewpoints will be present. Given that we show relationships do matter in judging appropriateness, we hope that future work will add diversity through new additions and data to study relationships. We will also release demographic information on annotators as a part of our dataset to help make potential biases more explicit and more easily addressed.

The annotators themselves were authors of the study and were compensated as a part of their normal work with a living wage. Due to the nature of our filtering, the vast majority of our content was not explicitly toxic. Nonetheless, some comments did contain objectionable messages, and annotators were provided guidance on how to seek self-care if the messages created distress.

With any new tool to identify offensive or abusive language comes a dual use by an adversarial actor to exploit that tool to find new ways to harass or abuse others while still ``abiding by the rules.'' Our work has shown that relationships are effective context (and features) for identifying previously-unrecognized inappropriateness. This new capability has the benefit of potentially recognizing more inappropriate messages before they reach their destination. However, some adversaries could still use our data and model to screen their own messages to find those that still are classified as appropriate (while being inappropriate in practice) to evade detection. Nevertheless, given the new ability to identify context-sensitive offensive messages---which we show can represent a substantial percentage of conversation (Section \ref{sec:pride})---we view the benefits as outweighing the risk.

\bibliography{anthology,custom}
\bibliographystyle{acl_natbib}

\appendix

\section{Annotation Details}
\label{app:annotation}

This section describes the details of the annotation process. Annotators were the authors of this paper and were compensated for their work as a part of their normal duties; no additional payments were provided.  

\begin{figure*}[t]
    \centering
    \includegraphics[width=\linewidth]{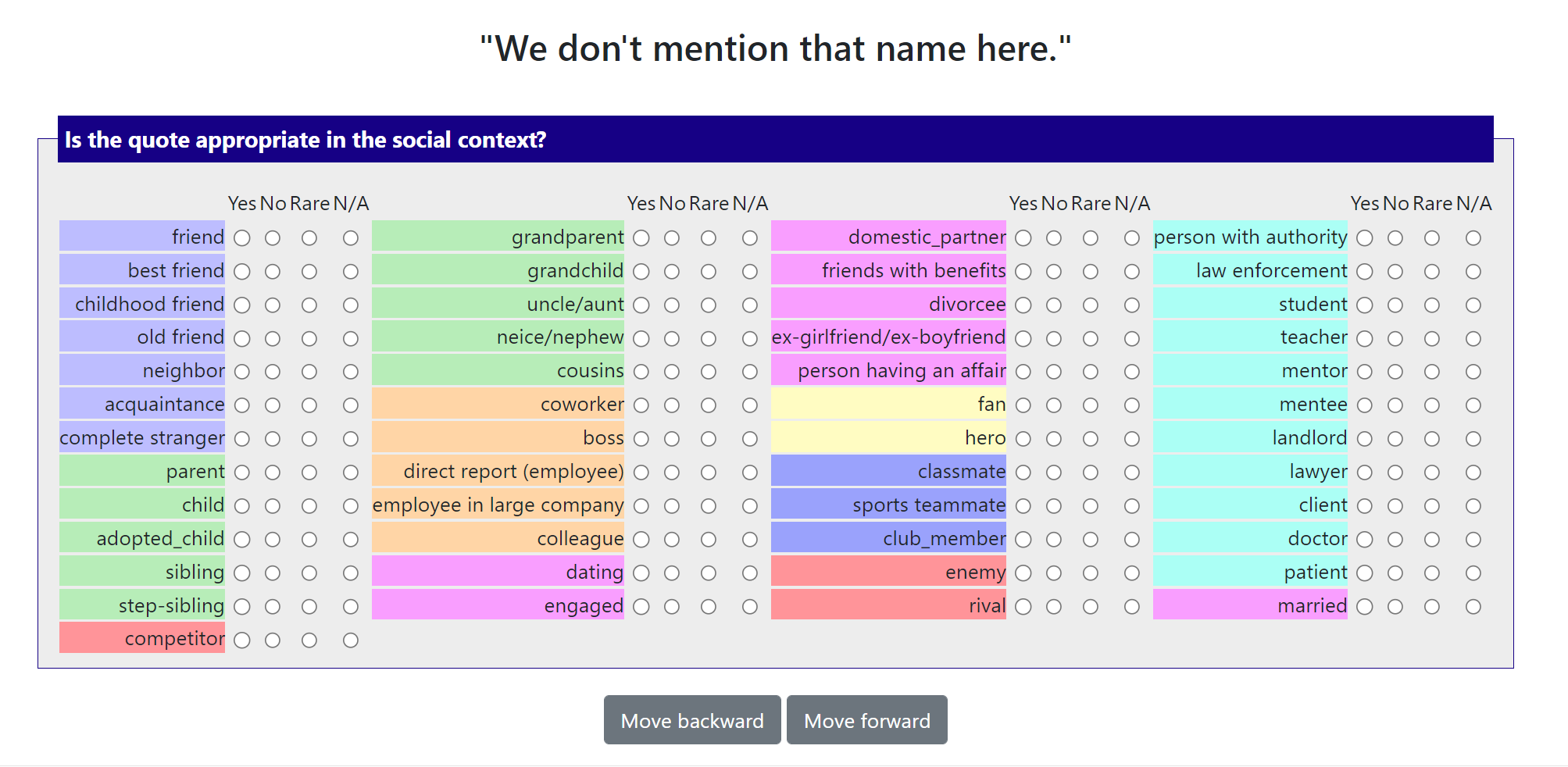}
    \caption{A screenshot of the annotation interface. Annotators were asked to first decide whether the message was plausible for a relationship and, if not, mark it as N/A. A ``Rare'' category was added for cognitive ease if an annotator thought there could be some plausible situation for the message, but this situation would be very rare. In practice, this option was rarely used and treated as N/A. Relationships were color-coded by folk category to help annotators annotate more easily.}
    \label{fig:annotation}
\end{figure*}

The annotation interface was designed using \textsc{Potato} \citep{pei2022potato}, shown in Figure~\ref{fig:annotation}, and was accessed through a browser, which  allowed annotators to start and stop their labeling at any time. Annotators were allowed to revise their annotations at any time. 

\begin{figure*}[t]
    \centering
    \includegraphics[width=\linewidth]{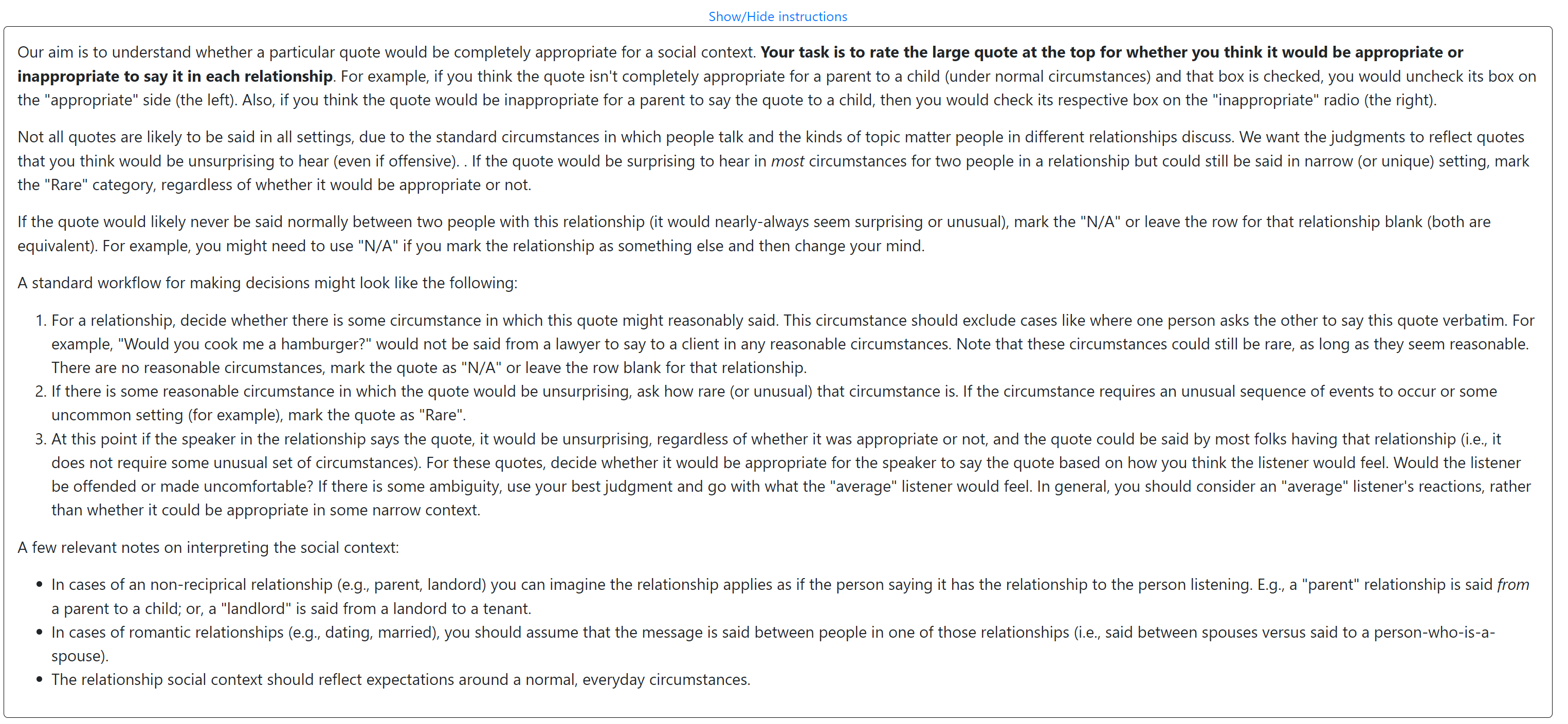}
    \caption{Screenshot of the Instructions provided to annotators during the annotation process. Instructions were available via a drop-down menu in the annotation interface, which was accessible through the entire process for their review.}
    \label{fig:annotation_instruction}
\end{figure*}

During annotation, annotators were presented with the message to be annotated and collapsible instructions for annotation. Figure~\ref{fig:annotation_instruction} shows the full written instructions shown to annotators. The instructions were refined through an iterative process throughout the project, and annotators regularly communicated about ambiguity. The instructions were designed to let the annotators know the intent of the study and the downstream tasks that data would be used for. 

\section{Conversation Classifier Details}
\label{app:conversational}

The conversational classifier was used during the initial data sampling phase to identify comments on Reddit that could plausibly have been said in a conversation. 
This classifier is intended only as a filter to improve data quality by reducing the number of non-conversation comments (e.g., those with Reddit formatting, long monologues, and comments written in a non-conversational register). We have two datasets of known conversations: 70,949 turns from the Empathetic dialogs data~\citep{rashkin2019empathetic} and 225,907 turns from the Cornell movie dataset~\citep{danescu2011chameleons} as positive examples of conversational messages. We then sample an equivalent number of 296,854 turns from a random sample of Reddit comments as non-conversational messages.  While some of these Reddit messages are likely conversational, this classification scheme is only a heuristic aimed at helping filter data.  A held-out set of 74,212 instances was used for evaluation, balanced between conversational and not.

A MiniLM classifier \citep{wang2020minilm} was trained using Huggingface Transformers~\citep{wolf2019transformer} for five epochs, keeping the model with the lowest training loss at any epoch; Epoch 5 was selected.  The model attained an F1 of 0.94 for the held-out data indicating it was accurate at distinguishing the conversational turns from the random sample of Reddit comments. We apply this classifier to 1,917,346 comments from Reddit during the month of February 2018 and identify 145,210 whose probability of being a conversation is $>$0.5. We retain these comments as potential comments to annotate in Phase 2 (Section~\ref{sec:phase2}).

\subsection{Computational resources}
All of our experiments were conducted on an Ubuntu 16.04.7 LTS machine installed with NVIDIA RTX A5000 and RTX A6000 GPUs having CUDA 11.3. The Python packages used in our experiments include Pytorch 1.17.0, Transformers 4.25.1, PEFT 0.3.0, OpenPrompt 1.0.1, pandas 1.1.4, spacy 3.3.2, and Sci-kit learn 1.2.0.

\subsection{Specification of LLMs }
The LLMs used in this paper were downloaded from \href{https://huggingface.co/}{huggingface.co}. The model and their parameter sizes are listed in Table \ref{table:model-param}. 
\begin{table}[th]
 \resizebox{0.48\textwidth}{!}{
\begin{tabular}{ c c ccc } 
Model & Label & No. parameters \\
\hline
 T5  \cite{2020t5}       &  t5-base & 220M \\ 
 GPT2 \cite{radford2019language}       & gpt2-medium &  355M   \\ 
 MiniLM \cite{wang2020minilm} & microsoft/MiniLM-L12-H384-uncased & 33M \\
DeBERTa-v3 \cite{he2021debertav3} & microsoft/deberta-v3-base & 86M\\
{FLAN-T5 \cite{chung2022scaling} } & {google/flan-t5-large} & {780M}\\
{FLAN-T5 \cite{chung2022scaling} } & {google/flan-t5-xl} & {3B}\\
\
\end{tabular}
}
\caption{A list of all pre-trained LLMs used in this study. The Label column corresponds to the label registered on the Hugging Face model repository.}
\label{table:model-param}
\end{table}

\subsection{Classifiers from Sklearn}
For the classification of politeness and condescension tasks, we used logistic regression from sklearn with the solver as `lbfgs' and max\_iter set 
 to 400.

\section{Phase 1 Classifier}
\label{app:pilot-classifier}

The phase-1 LLM classifier was trained using the pilot training data and the OpenPrompt framework. In this framework, we use a batch size of 4, the maximum sequence length was set to 256,  decoder\_max\_length=3, truncate\_method="head", and teacher\_forcing and predict\_eos\_token were set to default values. The prompt used for the model was framed as a yes/no question - "is it appropriate for PERSON1 to say QUOTE to PERSON2?".

\begin{table}[tb]
 \resizebox{0.48\textwidth}{!}{
\begin{tabular}{ c c ccc } 
Model & Type & Precision & Recall & F1 \\
\hline
 T5-base         & prompt-based & 0.67 & 0.61 & 0.64\\ 
 GPT2-med        & prompt-based & 0.71 & 0.57 & 0.63 \\ 
 random          & \textit{n/a} & 0.44 & 0.39 & 0.41 \\
\end{tabular}
}
\caption{Performance (Binary F1) at recognizing whether a message was  \textit{in}appropriate in a relationship context using the pilot training and test data.}
\label{table:pilot-model-performance}
\end{table}

\begin{table}[tb]
 \resizebox{0.48\textwidth}{!}{
\begin{tabular}{ c c ccc } 

 Model & Type & Precision & Recall & F1 \\
 \hline
 \textit{random} & \textit{n/a} & 0.43 & 0.33 & 0.36 \\  
 DeBERTa-v3 & supervised     & {0.702$\pm$0.20 }& {0.652 $\pm$ 0.037} & {0.676$\pm$0.027} \\
 MiniLM     & supervised     & {0.690 $\pm$ 0.016} & {0.713 $\pm$ 0.048} & {0.701 $\pm$ 0.021}\\
 T5-base    & prompt-based   & {0.710$\pm$0.075}   & {0.728 $\pm$ 0.026} & {0.723 $\pm$ 0.040} \\ 
 GPT2-med   & prompt-based   & {0.638 $\pm$ 0.043} & {0.720 $\pm$ 0.028} & {0.697 $\pm$ 0.012 } \\ 
flan-t5-large & prompt-based & {0.683 $\pm$ 0.019} & {0.726 $\pm$ 0.043} & {0.703 $\pm$ 0.014 } \\  
 flan-t5-xl  & prompt-base d & {0.717 $\pm$ 0.027} & {0.763 $\pm$ 0.072} & {0.740 $\pm$ 0.020} \\ 
 
 \hline
\end{tabular}
}
\caption{Performance of different trained models on the development dataset. Performance on the test set is reported in Table~\ref{tab:full-model-performance}.}
\label{table:model-dev-performance}
\end{table}

\begin{table}[tb]
 \resizebox{0.48\textwidth}{!}{
\begin{tabular}{ c c ccc } 

 Model & Type & Precision & Recall & F1 \\
\hline
 \textit{random} & \textit{n/a} & 0.44 & 0.37 & 0.40  \\
 Perspective API & \textit{n/a} & 0.42 & 0.097 & 0.16 \\ 
 DeBERTa-v3      & LM fine-tuning   & {0.658$\pm$0.019} & {0.66$\pm$0.010} & {0.659$\pm$ 0.014}\\
 MiniLM          & LM fine-tuning   & {0.615$\pm$0.035} & {0.705$\pm$0.023} & {0.656$\pm$0.017}\\
 T5-base         & prompt-based & {0.655 $\pm$ 0.018} & {0.683 $\pm$ 0.017} &  {0.669$\pm$0.012}\\ 
 GPT2-med        & prompt-based & {0.668 $\pm$ 0.008} & {0.650 $\pm$ 0.024} &  {0.665$\pm$0.018}\\ 
 flan-t5-large   & prompt-based & {0.626 $\pm$ 0.016} & {0.704 $\pm$ 0.056} & {0.661 $\pm$ 0.021}\\ 
 flan-t5-xl      & prompt-based & {0.666 $\pm$ 0.022} & {0.736 $\pm$ 0.041} & {0.698 $\pm$ 0.010} \\ 
\end{tabular}
}
\caption{Performance (Binary F1) at recognizing whether a message was  \textit{in}appropriate in a relationship context on the test set.}
\label{tab:full-model-performance}
\end{table}

\section{Additional Prompt-based Model Details}
\label{app:prompt-models}

We train \texttt{gpt2-base} and \texttt{t5-base} using the OpenPrompt framework. In this framework, we use a batch size of {16}, the maximum sequence length was set to 256,  decoder\_max\_length=3, truncate\_method="head", and teacher\_forcing and predict\_eos\_token were set to default values. The model was trained using early stopping and the AdamW optimizer with a learning rate set to 1e-4. The different prompts that we used before finalizing the prompt as {"Is it appropriate for PERSON1 to say "QUOTE" to PERSON2?, "yes" or "no"?} are reported in table \ref{tab:prompt}.

We train the \texttt{flan-t5-large} and \texttt{flan-t5-xl} models using the PEFT library. Models were trained with a batch size of 96 and 32, respectively. Both models used a maximum sequence length of 192 and learning rate of 1e-2 with AdamW, using all other default library parameters. The model was trained for 20 epochs, keeping the best-performing model by binary F1 on the development dataset for each seed.

\begin{table*}[t]
\resizebox{\textwidth}{!}{
\begin{tabular}{ clllc } 
Model & Prompt & Verbalisation & Binary-F1 (Mean) & Standard Deviation \\
 \hline
 \textbf{\texttt{t5-base} }& \textbf{A PERSON1 saying ``QUOTE" to PERSON2 would be \{mask\}} & \textbf{appropriate/inappropriate }& \textbf{0.669} & \textbf{0.012} \\
 \texttt{t5-base} & Is it appropriate for PERSON1 to say ``QUOTE" to PERSON2, ``yes" or ``no"? \{mask\}& yes/no & 0.657 & 0.027\\
 \texttt{t5-base}& Would it be appropriate for PERSON1 to say ``QUOTE" to PERSON2, ``yes" or ``no"? \{mask\}& yes/no & 0.661 & 0.025\\
 \texttt{t5-base} & Would it be ``more" or ``less" appropriate for PERSON1 to say ``QUOTE" to PERSON2? \{mask\} & less/more & \textbf{0.669} & \textbf{0.021}\\
 \texttt{t5-base}& If PERSON1 says ``QUOTE" to PERSON2, would it be ``more" or ``less" appropriate? \{mask\}& less/more & 0.660 & 0.019\\
 \hline
\textbf{\texttt{gpt2-med}} & \textbf{A PERSON1 saying ``QUOTE" to PERSON2 would be} & \textbf{appropriate/inappropriate} & \textbf{0.665} &\textbf{ 0.018} \\
\texttt{gpt2-med} & Is it appropriate for PERSON1 to say ``QUOTE" to PERSON2, ``yes" or ``no"? & yes/no & 0.612 & 0.036\\
\texttt{gpt2-med} & Would it be appropriate for PERSON1 to say ``QUOTE" to PERSON2, ``yes" or ``no"? \{mask\} & yes/no & 0.632 & 0.009\\
\texttt{gpt2-med} & Would it be ``more" or ``less" appropriate for PERSON1 to say ``QUOTE" to PERSON2? \{mask\}& less/more &0.630 & 0.021 \\
\texttt{gpt2-med}& If PERSON1 says ``QUOTE" to PERSON2, would it be ``more" or ``less" appropriate? \{mask\} & less/more &0.652 & 0.02\\

\end{tabular}
}
\caption{Binary F1 score (test) for various prompts used with the LLMs in the Openprompt Framework } 
\label{tab:prompt}
\end{table*}

\section{Additional Results}
\label{app:more-results}

\subsection{Development Set Performance}

The performance of the different models on the development dataset is reported in  Table~\ref{table:model-dev-performance} and performance on the test set with standard errors is reported in Table~\ref{tab:full-model-performance}. 

\subsection{Analysis of Relationship Predictions}

The data annotation process showed clear associations between pairs of relationships in terms of how often a message would be appropriate (Figure~\ref{fig:app-heatmap}). However, the training data for that figure only includes annotations on relationships annotators selected. What structure or regularity might we see from analyzing similarities between all our relationships through model predictions?

As a qualitative experiment, we use the \texttt{flan-t5-xl} model to label the subset of the PRIDE dataset (Section~\ref{sec:pride}) for the appropriateness of all 49 relationships in our training data. This produces a binary matrix of 49 $\times$ 47,801. We use PCA to capture regularity and then project relationships onto a 2D visualization using t-SNE \cite{JMLR:v9:vandermaaten08a}, which is aimed at preserving local similarity in the spatial arrangement.
If model predictions are capturing shared norms, we view t-SNE as potentially more useful than a PCA projection, as we want to visualize which relationships with similar judgments as being nearby (what t-SNE does) rather than optimizing the visualization to  the global structure of distances (what PCA does).
The t-SNE projection was designed using guidance from \citet{wattenberg2016use}; a perplexity of 40 was used.

\begin{figure*}[tb]
    \centering
    \includegraphics[width=0.75\linewidth]{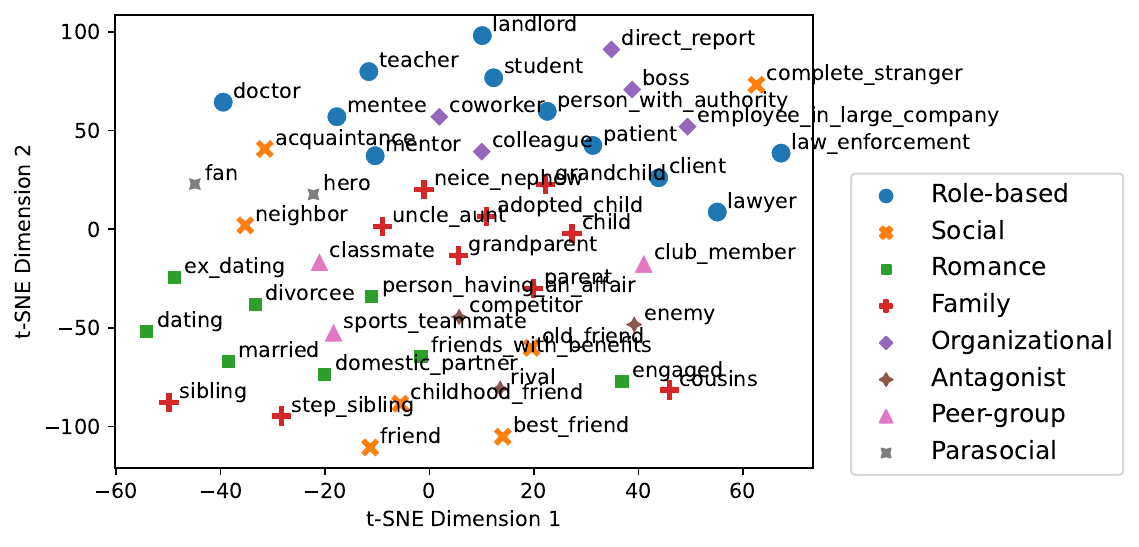}
    \caption{A plot of relationships, projected using t-SNE from a PCA of their appropriateness judgments across the 47,801 messages we use from the PRIDE dataset. }
    \label{fig:tsne}
\end{figure*}

The resulting visualization, shown in \fref{fig:tsne}, captures expected regularity. While the projection is only a visual tool, and aspects such as distance are not meaningful in t-SNE visualizations, the grouping and neighbors suggest the model is sensitive to power/status and social distance in how it decides appropriateness based on the relationship. 

\subsection{Per Relationship Results}

Table \ref{tab:per-rel-performance} shows the peformance of the \texttt{flan-t5-xl} model on the test set, broken down by relationship

\begin{table*}[t]
\centering
\resizebox{0.9\textwidth}{!}{
\begin{tabular}{lrrrrlr}
Relationship & Precision & Recall & F1 & \# Training Examples & Category & \% Offensive \\
\hline
hero & 1.00 & 1.00 & 1.00 & 36 & Parasocial & 0.19 \\
doctor & 1.00 & 0.88 & 0.93 & 85 & Role Based & 0.47 \\
student & 0.83 & 1.00 & 0.91 & 115 & Role Based & 0.51 \\
client & 0.83 & 1.00 & 0.91 & 41 & Role Based & 0.37 \\
boss & 0.82 & 0.97 & 0.89 & 330 & Organizational & 0.72 \\
patient & 0.80 & 1.00 & 0.89 & 40 & Role Based & 0.40 \\
fan & 0.80 & 1.00 & 0.89 & 34 & Parasocial & 0.15 \\
direct report & 0.81 & 0.98 & 0.89 & 278 & Organizational & 0.68 \\
person with authority & 0.82 & 0.95 & 0.88 & 114 & Role Based & 0.61 \\
teacher & 0.83 & 0.92 & 0.87 & 217 & Role Based & 0.62 \\
lawyer & 0.89 & 0.80 & 0.84 & 76 & Role Based & 0.54 \\
landlord & 0.73 & 0.92 & 0.81 & 110 & Role Based & 0.67 \\
employee in large company & 0.70 & 0.92 & 0.80 & 230 & Organizational & 0.59 \\
uncle aunt & 0.72 & 0.88 & 0.79 & 202 & Family & 0.36 \\
complete stranger & 0.70 & 0.91 & 0.79 & 234 & Social & 0.64 \\
child & 0.78 & 0.78 & 0.78 & 172 & Family & 0.50 \\
law enforcement & 0.67 & 0.92 & 0.77 & 140 & Role Based & 0.63 \\
mentee & 0.65 & 0.94 & 0.77 & 111 & Role Based & 0.48 \\
colleague & 0.68 & 0.87 & 0.76 & 245 & Organizational & 0.56 \\
grandchild & 0.73 & 0.79 & 0.76 & 185 & Family & 0.45 \\
niece/nephew & 0.71 & 0.80 & 0.75 & 125 & Family & 0.54 \\
adopted child & 0.71 & 0.80 & 0.75 & 170 & Family & 0.44 \\
acquaintance & 0.68 & 0.83 & 0.75 & 193 & Social & 0.56 \\
coworker & 0.63 & 0.90 & 0.75 & 291 & Organizational & 0.52 \\
neighbor & 0.64 & 0.88 & 0.74 & 217 & Social & 0.48 \\
parent & 0.82 & 0.68 & 0.74 & 296 & Family & 0.27 \\
grandparent & 0.71 & 0.77 & 0.74 & 211 & Family & 0.39 \\
competitor & 0.50 & 1.00 & 0.67 & 71 & Antagonist & 0.18 \\
enemy & 0.60 & 0.75 & 0.67 & 76 & Antagonist & 0.18 \\
mentor & 0.52 & 0.81 & 0.63 & 157 & Role Based & 0.45 \\
club member & 0.53 & 0.75 & 0.62 & 125 & Peer group & 0.29 \\
ex dating & 0.53 & 0.71 & 0.61 & 222 & Romance & 0.27 \\
divorcee & 0.59 & 0.62 & 0.61 & 208 & Romance & 0.24 \\
domestic partner & 0.50 & 0.73 & 0.59 & 211 & Romance & 0.19 \\
sports teammate & 0.53 & 0.62 & 0.57 & 156 & Peer group & 0.31 \\
classmate & 0.47 & 0.69 & 0.56 & 140 & Peer group & 0.34 \\
married & 0.47 & 0.70 & 0.56 & 310 & Romance & 0.21 \\
friends with benefits & 0.47 & 0.64 & 0.54 & 235 & Romance & 0.20 \\
person having an affair & 0.44 & 0.67 & 0.53 & 186 & Romance & 0.19 \\
engaged & 0.40 & 0.73 & 0.52 & 287 & Romance & 0.21 \\
dating & 0.47 & 0.56 & 0.51 & 288 & Romance & 0.23 \\
rival & 0.40 & 0.67 & 0.50 & 64 & Antagonist & 0.11 \\
sibling & 0.20 & 0.50 & 0.29 & 241 & Family & 0.10 \\
old friend & 0.29 & 0.24 & 0.26 & 269 & Social & 0.22 \\
step sibling & 0.15 & 0.50 & 0.24 & 194 & Family & 0.10 \\
cousins & 0.13 & 0.43 & 0.20 & 207 & Family & 0.19 \\
best friend & 0.14 & 0.25 & 0.18 & 364 & Social & 0.06 \\
friend & 0.12 & 0.14 & 0.13 & 335 & Social & 0.10 \\
childhood friend & 0.15 & 0.11 & 0.13 & 263 & Social & 0.23 \\
\end{tabular}
}
\caption{Performance of the \texttt{flan-t5-xl} model on the test set per relationship type, ordered by binary F1. }
\label{tab:per-rel-performance}

\end{table*}

\section{Additional Experimental Setup Details for Identifying Subtly Offensive Messages}
\label{app:subtle}

For experiments with both the TalkDown corpus \cite{wang2019talkdown} and Stanford Politeness Corpus \cite{danescu2013computational}, the trained \texttt{flan-t5-xl} model was used in a zero-shot setting with no additional training. For the Politeness corpus, politeness ratings are made at the utterance level, outside of any dialog context. As a result, the existing prompt was used (\sref{sec:identifying}, Experimental Setup) to assess relationship-specific appropriateness. 

Two modifications were necessary for the TalkDown corpus. First, the TalkDown corpus's data is rated at the \textit{turn} level, with condescension judgments based on the interpretation of a reply to a specific piece of quoted text. \citet{wang2019talkdown} note that incorporating both the quote and reply into the input resulted in better performance. Therefore, we modify our initial prompt slightly as follows: 
``Rate whether it is inappropriate for message A to be said in response to the message B in the specified social setting: \textbackslash n A: {quoted text} \textbackslash n B: {reply text} \textbackslash n setting: {relationship description} \textbackslash n answer (yes or no):''. 
Since the \texttt{flan-t5-xl} model was trained specifically for instruction following  \cite{chung2022scaling}, we expected the model to generate similar outputs as our original prompt.
Second, some of the quoted and reply text in TalkDown can be quite long (hundreds of words). Since the adapted prompt contains both quote and reply, we use an flexible truncation process to maximize the content that can still fit within the maximum input token sequence length (196). First, quoted text over 50 tokens is truncated to the first 50, using the \texttt{flan-t5-xl} tokenizer to segment words. Then, if the full input (with prompt instructions) still exceeds the maximum input length, we truncate both the quoted text and reply evenly, still keeping at least the first then 10 tokens of each.

\end{document}